\documentclass[11pt]{article}

\usepackage[final]{acl}

\usepackage{times}
\usepackage{latexsym}
\usepackage{marvosym}
\usepackage{makecell}

\usepackage[T1]{fontenc}

\usepackage[utf8]{inputenc}

\usepackage{microtype}

\usepackage{inconsolata}
\usepackage{amsmath}
\usepackage{enumitem}

\usepackage{booktabs}
\usepackage{graphicx}
\usepackage{multirow}
\usepackage{graphicx}
\usepackage{pifont} 
\usepackage[dvipsnames]{xcolor}
\usepackage{colortbl}
\usepackage{hyperref}
\usepackage[capitalise]{cleveref}

\usepackage{algorithm}
\usepackage{algpseudocode}
\usepackage{amssymb}
\usepackage[most]{tcolorbox}
\tcbuselibrary{breakable}
\usepackage{siunitx}

\newcommand{\cmark}{\ding{51}}%
\newcommand{\xmark}{\ding{55}}%
\newcommand{\method}{Trainee-Bench}
\title{The Agent's First Day: Benchmarking Learning, Exploration, and Scheduling in the Workplace Scenarios}

\author{
    \textbf{Daocheng Fu\textsuperscript{1,2,$^{\dag}$}},
    \textbf{Jianbiao Mei\textsuperscript{3,2,$^{\dag}$}},
    \textbf{Rong Wu\textsuperscript{3,2,$^{\dag}$}},
    \textbf{Xuemeng Yang\textsuperscript{2,$^{\dag}$}},
    \\
    \textbf{Jia Xu\textsuperscript{2}},
    \textbf{Ding Wang\textsuperscript{2}},
    \textbf{Pinlong Cai\textsuperscript{2}},
    \textbf{Yong Liu\textsuperscript{3,$^{\textrm{\Letter}}$}},
    \textbf{Licheng Wen\textsuperscript{2,4,5,$^{\textrm{\Letter}}$}},
    \textbf{Botian Shi\textsuperscript{2,$^{\textrm{\Letter}}$}}
    \\
    \textsuperscript{1}Fudan University,
    \textsuperscript{2}Shanghai AI Laboratory,
    \textsuperscript{3}Zhejiang University,
    \\
    \textsuperscript{4}Shanghai Innovation Institute,
    \textsuperscript{5}Shanghai Jiao Tong University
    \\
    \small{
        $^{\dag}$ Equal contribution, $^{\textrm{\Letter}}$ Corresponding authors.
    }
}

\begin{document}
\maketitle

\begin{abstract}
The rapid evolution of Large Language Models (LLMs) has advanced workflow automation; however, existing research mainly targets performance upper bounds in static environments, overlooking robustness for stochastic real-world deployment. We identify three key challenges: dynamic task scheduling, active exploration under uncertainty, and continuous learning from experience. To bridge this gap, we introduce \method{}, a dynamic evaluation environment that simulates a "trainee" agent continuously exploring a novel setting. Unlike traditional benchmarks, \method{} evaluates agents along three dimensions: (1) context-aware scheduling for streaming tasks with varying priorities; (2) prudent information acquisition to reduce hallucination via active exploration; and (3) continuous evolution by distilling generalized strategies from rule-based, dynamically generated tasks. Experiments show that cutting-edge agents have significant deficiencies in dynamic environments, especially in active exploration and continual learning. Our work establishes a framework for assessing agent reliability, shifting evaluation from static tests to realistic, production-oriented scenarios. Our codes are available at \href{https://github.com/KnowledgeXLab/EvoEnv}{https://github.com/KnowledgeXLab/EvoEnv}
\end{abstract}
\section{Introduction}

The rapid evolution of Large Language Models (LLMs) has significantly advanced complex workflow automation~\cite{guo2024large, li2024survey}. However, while existing systems facilitate tool utilization, research predominantly focuses on performance \textit{upper bounds} in controlled environments~\cite{gottweis2025towards, internagentteam2025internagentagentscientist}. This paradigm often neglects the robustness required for authentic, stochastic environment settings. Unlike sterile experiments that ignore environmental noise, real-world deployment involves streaming, randomized tasks. Consequently, agents must not only orchestrate scheduling with awareness of environmental dynamics but also ensure reliability, effectively bridging the gap between laboratory settings and production needs~\cite{fu2025re,yang2025learning}.

\begin{figure}[t!]
    \centering
    \includegraphics[width=\linewidth]{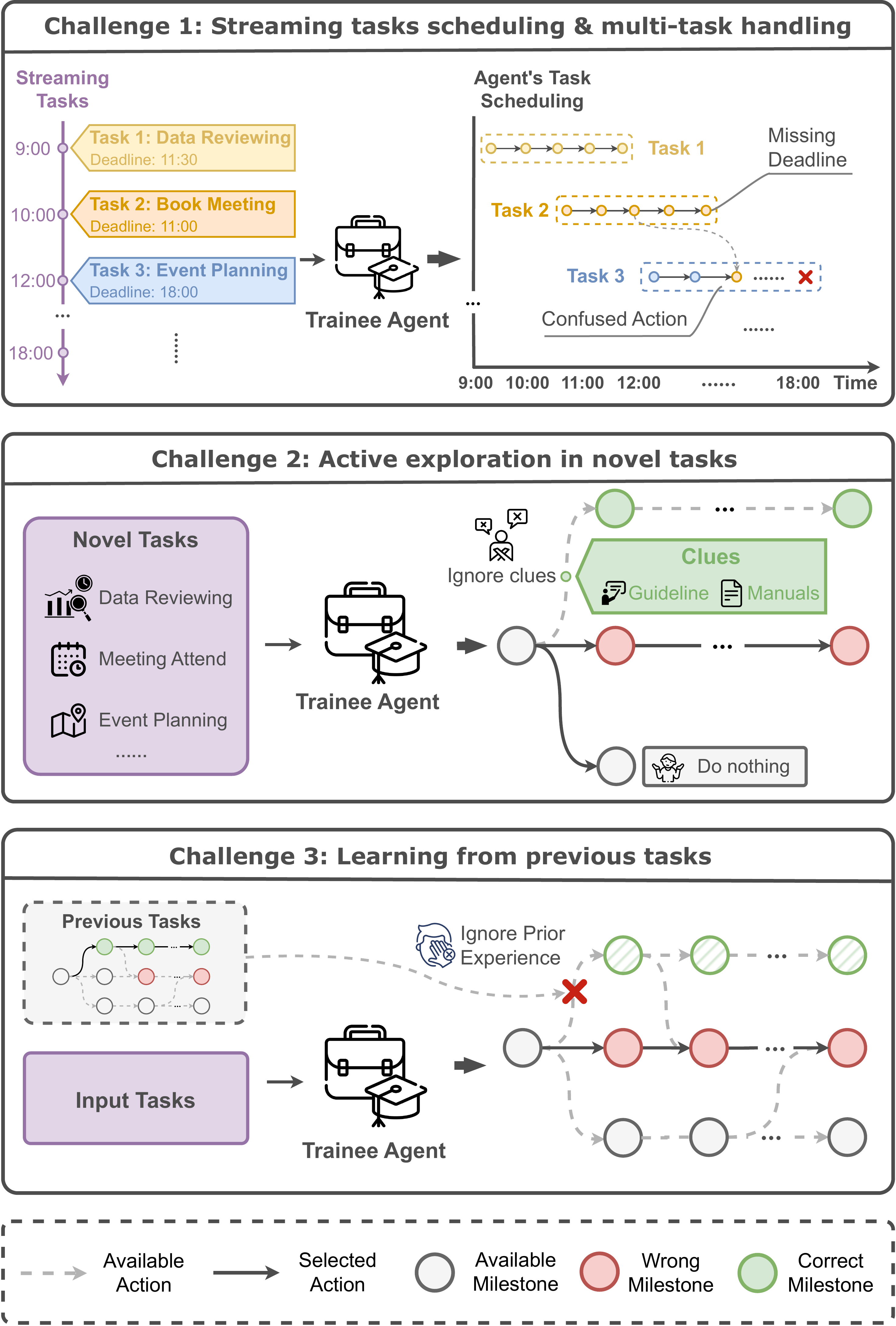}
    \caption{Overview of challenges in current agent systems: (1) effective scheduling and multi-task planning for streaming task inputs; (2) the ability to suspect unsolvable tasks and solicit guidance through active exploration; and (3) robust experience summarization, retrieval, and utilization to enhance performance stability.}
    \label{fig:problem_illustration}
    \vspace{-1em}
\end{figure}

\begin{table*}[t]
\centering
\small
\setlength{\tabcolsep}{2pt}
\resizebox{\linewidth}{!}{%
\begin{tabular}{l c c c c}
\toprule
\textbf{Benchmark} & 
\textbf{Multi-App Interaction} & 
\textbf{Checkpoint Feedback} & 
\textbf{Partial Observability} & 
\textbf{Dynamic Configuration} \\
\midrule
GAIA-2~\cite{froger2025scaling} & \cmark & \xmark & \xmark & \xmark \\
TheAgentCompany~\cite{xu2024theagentcompany} & \cmark & \cmark & \xmark & \xmark \\
Tool Bench~\cite{qin2024tool} & \cmark & \xmark & \xmark & \xmark \\
StableToolBench~\cite{guo2024stabletoolbench} & \cmark & \xmark & \xmark & \xmark \\
Toolathon~\cite{li2025tool} & \cmark & \xmark & \xmark & \xmark \\
$\tau$-bench~\cite{yao2024tau} & \xmark & \xmark & \cmark & \xmark \\
\midrule
\rowcolor{gray!15} \textbf{{\method} (Ours)} & \cmark & \cmark & \cmark & \cmark \\
\bottomrule
\end{tabular}%
}
\caption{Comparison of different agent benchmarks. ``Multi-App Interaction'' : support for cross-application workflows. ``Checkpoint Feedback'': the availability of intermediate evaluation milestones beyond final success rates. ``Partial Observability'': agents must actively explore hidden states rather than operating on full information. ``Dynamic Configuration'': environments with randomized, evolving parameters and dynamic composite scenarios.}
\vspace{-1em}
\label{tab:benchmarks}
\end{table*}

As illustrated in~\cref{fig:problem_illustration}, shifting from static setups to realistic interaction scenarios introduces three primary challenges:
(1) \textbf{Dynamic task scheduling and context management.} Agents must perform rational temporal planning for streaming tasks to meet deadlines while maintaining context awareness, thereby deciding to mitigate cross-task interference.
(2) \textbf{Active exploration in novel tasks.} Instead of hallucinating actions in uncertain scenarios, a reliable agent should exercise prudence by actively exploring or soliciting assistance to acquire necessary clues~\cite{TurtleBench,qiu2025quantifying, arora2025healthbench}.
(3) \textbf{Learning from previous tasks.} To ensure long-term stability, agents must learn from prior experiences to prevent the recurrence of historical errors in subsequent scenarios~\cite{zhou2025memento,silver2025welcome,wu2025evolver,gao2025survey}.

Although numerous benchmarks exist (~\cref{tab:benchmarks}), they fall short in simulating the dynamic nature of realistic environments and lack mechanisms to evaluate active exploration and continual learning capabilities.
To address this, we introduce \method{}, a dynamic evaluation environment simulating an ``trainee'' agent continuously exploring a novel setting. \method{} is designed to assess the three aforementioned capabilities of an agent rigorously:
First, regarding \textbf{task execution}, \method{} presents streaming tasks with distinct priorities. This demands superior scheduling to maintain context awareness and mitigate inter-task interference. 
Second, concerning \textbf{information acquisition}, critical clues are initially concealed. Agents must demonstrate prudence by actively exploring or soliciting help rather than engaging in hallucinated actions. 
Finally, in terms of \textbf{continuous evolution}, tasks are dynamically generated based on rules rather than fixed datasets. This compels agents to distill generalized strategies from prior tasks instead of rote-memorizing instances, effectively preventing repeated errors.

We conduct extensive experiments on \method{}, revealing that state-of-the-art (SOTA) agents exhibit significant room for improvement in dynamic environments, particularly regarding active exploration and continual learning. Our contributions are summarized as follows:
\begin{itemize}[leftmargin=1em, itemsep=0pt, topsep=2pt]
    \item Proposing a novel benchmarking paradigm that shifts from static, laboratory evaluations to stochastic, production-oriented environments;
    \item Establishing a comprehensive framework to rigorously assess temporal scheduling, prudent decision-making under uncertainty, and long-term strategic evolution of MLLM agents;
    \item Through extensive experimentation, we characterize the reliability gap of current agents, highlighting their deficiencies in handling concealed clues and distilling generalized experiences.
\end{itemize}


\section{\method}
We introduce \method{}, a benchmark designed to simulate authentic workplace environments through a series of human-crafted meta-tasks. In this framework, agents are evaluated as ``corporate interns'' navigating realistic company routines, such as attending meetings or reviewing data. The benchmark rigorously assesses three core competencies: dynamic multi-tasking, proactive exploration under uncertainty, and continuous self-evolution. 
To construct \method{}, we adopt a bottom-up design strategy that evaluates agent capabilities ranging from atomic skills to holistic workflows. We first develop a suite of rule-based \textit{meta-tasks} to serve as foundational templates. These templates can be instantiated into numerous specific task instances, which are then flexibly orchestrated into complex, composite scenarios with explicit temporal constraints. Finally, we implement an automated verification mechanism to ensure a rigorous and objective assessment of agent performance in these sophisticated environments. Detailed implementations are elaborated in the following sections.


\subsection{Preliminaries}
In \method{}, we formalize the interaction environment as a dynamic state transition system. At any given time step $t$, the environmental configuration is represented by a state $m_t \in \mathcal{M}$, which encapsulates the real-time status of the agent and various environmental entities, such as non-player characters (NPCs), files, and databases:
\begin{equation}
    m_t = \{s_{\text{agent}}, s_{\text{NPC}}, s_{\text{file}}, s_{\text{data}}\}
\end{equation}
The agent interacts with the environment by invoking tools to execute actions $a_t \in \mathcal{A}$. These actions trigger state transitions defined by the function $\mathcal{T}: \mathcal{M} \times \mathcal{A} \rightarrow \mathcal{M}$, defined as $\mathcal{T}(m_t, a_t) = m_{t+1}$. This formalism establishes the mathematical foundation for both the procedural generation of individual meta-tasks and the dynamic orchestration of concurrent, multi-threaded workflows.


\begin{figure}
    \centering
    \includegraphics[width=1.07\linewidth]{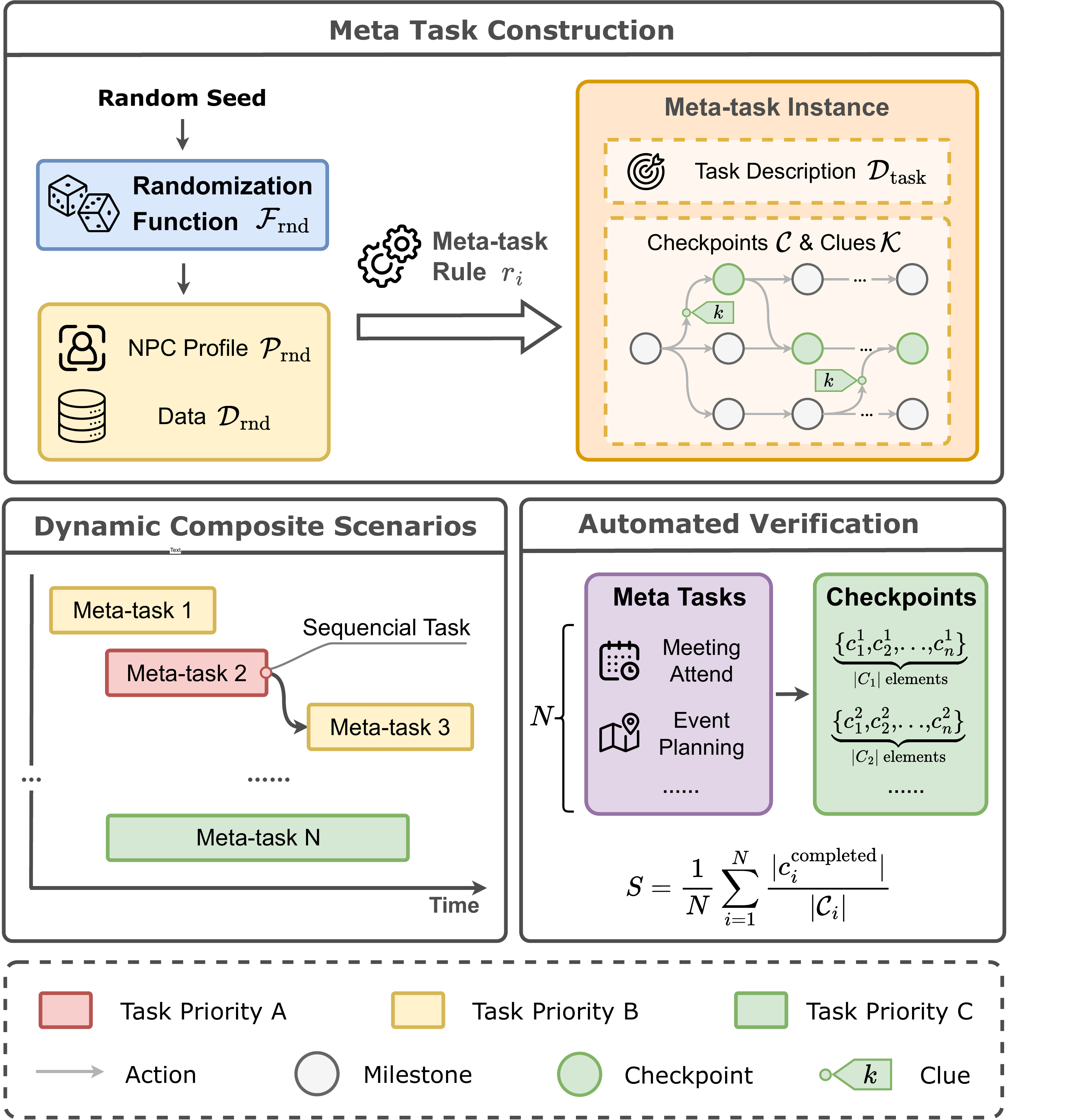}
    \caption{Overview of \method{} construction: (1) Unique task instances generated via random parameters. (2) Temporal composition of instances into task streams. (3) Automated verification of execution results.}
    
    \label{fig:construction_pipeline}
    \vspace{-10pt}
\end{figure}




\subsection{Meta-Task Design}
\label{sec:instantiation}

To prevent agents from over-fitting to static datasets, \method{} generates dynamic task instances derived from a library of human-crafted \textbf{Meta-tasks}, denoted as $\mathcal{R}$. Each $r_i \in \mathcal{R}$ represents an abstract logical template (e.g., ``Transaction Auditing'') that can spawn an infinite variety of concrete task instances.

\paragraph{Dynamic Task Instantiation.}
The instantiation process begins with a randomization engine $\mathcal{F}_{\text{rnd}}$. Controlled by a random seed, this engine synthesizes stochastic environment variables, specifically NPC profiles $\mathcal{P}_{\text{rnd}}$ and environmental data $\mathcal{D}_{\text{rnd}}$:
\begin{equation}
    \{\mathcal{P}_{\text{rnd}}, \mathcal{D}_{\text{rnd}}\} = \mathcal{F}_{\text{rnd}}(\text{seed})
\end{equation}
These variables introduce realistic uncertainty, such as randomized sender names, file paths, or conflicting schedules. Then, the selected meta-task rule $r_i$ integrates these variables into its predefined logic to yield a task triple:
\begin{equation}
    \{\mathcal{P}_{\text{rnd}}, \mathcal{D}_{\text{rnd}}\} \xrightarrow{r_i} \{\mathcal{D}_{\text{task}}, \mathcal{C}, \mathcal{K}\},
\end{equation}
where $\mathcal{D}_{\text{task}}$ represents a unique natural language objective, $\mathcal{C}$ denotes a set of verifiable checkpoints for evaluation, and $\mathcal{K}$ comprises the latent clues required for task completion. By varying the seed, a single meta-task $r_i$ can generate multiple distinct task triples. This one-to-many mapping ensures that agents must employ general reasoning and actively explore latent clues rather than relying on rote memorization.

\paragraph{Task Diversity and Coverage.}
The decoupling of logical rules $r_i$ from stochastic data $\{\mathcal{P}_{\text{rnd}}, \mathcal{D}_{\text{rnd}}\}$ enables a systematic and scalable expansion of the benchmark's coverage along two dimensions:
\begin{itemize}[leftmargin=1em, itemsep=0pt, topsep=2pt]
    \item \textbf{Task Dimensions:} To assess a broad spectrum of cognitive demands rather than isolated skills, our rule library $\mathcal{R}$ comprises 181 meta-tasks categorized into four functional domains: (i) \textit{information synthesis and analysis} (e.g., transaction auditing), (ii) \textit{time management and scheduling} (e.g., meeting attending), (iii) \textit{proactive inquiry and monitoring} (e.g., automated website monitoring), and (iv) \textit{strategic modeling and optimization} (e.g., advertising campaign planning).
    \item \textbf{Stochastic Robustness:} The randomization function $\mathcal{F}_{\text{rnd}}$ ensures that even for a single rule $r_i$, the resulting environment remains unpredictable. By varying user personas, file system hierarchies, and numerical distributions within $\mathcal{D}_{\text{rnd}}$, \method{} prevents agents from memorizing specific environment layouts or hard-coded solutions, forcing them to generalize across diverse and unfamiliar operational contexts.
\end{itemize}

\paragraph{Partial Observability.}
A defining characteristic of \method{} is the deliberate enforcement of partial observability to simulate real-world ambiguity. We introduce an information gap between the agent's initial knowledge and the required state information. Specifically, while the agent is provided with the task objective $\mathcal{D}_{\text{task}}$, the set of essential clues $\mathcal{K} = \{k_1, \dots, k_n\}$—such as technical manuals, access passwords, or implicit verbal instructions—is strictly withheld from the initial prompt. Consequently, the agent cannot complete the task by simply following the starting instructions. Instead, it must engage in proactive exploration by navigating file systems or conducting multi-turn dialogues with NPCs to uncover the latent clues embedded in the environment. This mechanism shifts the agent's role from a passive executor to a proactive problem-solver that must synthesize fragmented information under uncertainty.

\subsection{Dynamic Composite Scenarios}
To simulate a realistic, multi-threaded scenario, we transcend the execution of isolated rules $r_i$ by synthesizing multiple meta-tasks into \emph{dynamic composite scenarios}.
Formally, a scenario is constructed by stochastically assembling a subset of generated tasks and distributing them along a continuous timeline, characterized by distinct deadlines. This composition introduces three critical structural constraints regarding the interaction:

\paragraph{Conflict Resolution:} During the aggregation of meta-tasks, conflicts may arise, such as overlapping entity identifiers (e.g., character names or filenames) or a single non-player character (NPC) acting as a custodian for clues $k_i^1$ and $k_j^2$ from disparate tasks. To resolve such conflicts, we enforce a uniqueness constraint on all entity demarcations within the meta-task rules, ensuring that specific names remain distinct across the composite scenario. Furthermore, in instances where a single NPC safeguards multiple clues, we explicitly define the conditional triggers required to elicit each specific clue\footnote{Detailed cases illustrating multi-clue NPC interactions and agent-NPC dialogues are provided in the~\cref{sec:clues_keeping}}, thereby mitigating potential information ambiguity.
    
\paragraph{Temporal Prioritization:} We introduce time-critical meta-tasks (e.g., attending a scheduled meeting) that impose rigid temporal constraints. These high-priority interrupts function as \textit{preemptive signals}, mandating that the agent suspend its current workflow, perform a context switch to address the immediate requirement, and subsequently resume the interrupted process. This mechanism rigorously evaluates the agent's capacity for dynamic priority scheduling and long-term memory management.

\paragraph{Inter-task Dependencies:} We engineer sequential dependencies where upstream tasks function as informational prerequisites for downstream objectives. In these scenarios, critical task parameters are only revealed during the execution of a preceding task—for instance, a new assignment may be "published" only during an ongoing meeting. This interdependent architecture increases temporal uncertainty and compels the agent to adaptively update its plan based on real-time environmental observations rather than relying on static initial instructions.

\begin{figure}[t]
    \centering
    \includegraphics[width=0.92\linewidth]{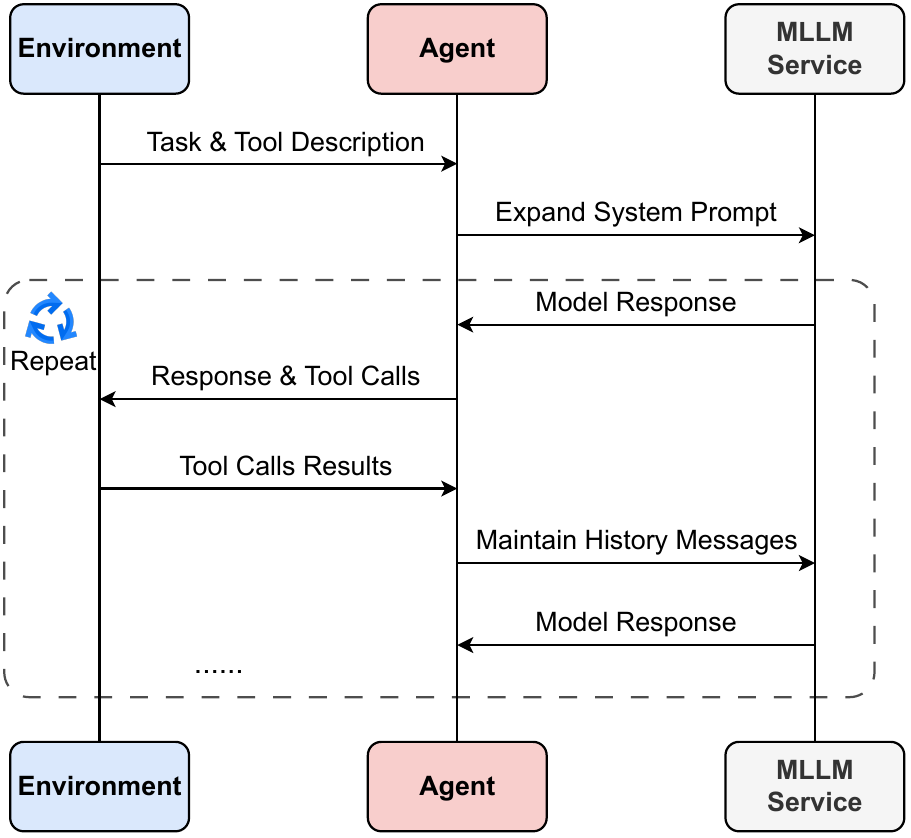}
    \caption{Formulation for the interaction among the environment, agent, and MLLM service.}
    \label{fig:task_formulation}
    \vspace{-10pt}
\end{figure}

\subsection{Automated Verification Mechanism}
To quantify performance in these complex scenarios, we leverage the set of checkpoints $\mathcal{C} = \{c_1, \dots, c_n\}$ derived during task construction. Embedded within each meta-task, these checkpoints facilitate a granular assessment of the agent's overall progress and completion quality. 
The final scenario score $S$ is derived by normalizing the completion ratio of checkpoints across all constituent meta-tasks:

\begin{equation}
   S = \frac{1}{N} \sum_{i=1}^{N} \frac{|c^{\text{completed}}_i|}{|\mathcal{C}_i|}
\end{equation} \label{eq:score}
where $N$ is the number of meta-tasks, $|\mathcal{C}_i|$ is the total number of checkpoints for task $i$, and $|c^{\text{completed}}_i|$ is the count of successfully passed checkpoints. Furthermore, each checkpoint triggers detailed natural language feedback, providing structured experiential data that empowers agents to perform retrospective analysis and continuous learning.

\subsection{Interaction Protocol}
\label{sec:interaction_protocol}
Building upon the constructed scenarios, we define an interaction protocol that bridges the environment and the agent, as depicted in Fig. \ref{fig:task_formulation}. This workflow comprises three distinct entities: the \emph{Environment}, the \emph{Agent}, and the \emph{MLLM Service}. Upon initialization with task descriptions and tool definitions provided by the environment, the agent constructs its system prompt and engages in an iterative cycle: maintaining conversation history, querying the MLLM Service, and parsing model responses into executable tool calls. We implement a standardized pipeline to orchestrate this Agent-Environment interaction loop, while abstracting three key components, i.e., \emph{system prompt construction}, \emph{history maintenance}, and \emph{MLLM service invocation}, as customizable interfaces. By enforcing a unified interaction protocol, \method{} provides a modular and flexible framework that supports diverse agent architectures and LLM backends.

\subsection{Benchmark Design Characteristics}
While conventional benchmarks often rely on isolated, static tasks that fail to capture the inherent complexity of real-world workflows, \method{} provides a high-fidelity workplace simulation by integrating four critical operational dimensions (see Table~\ref{tab:benchmarks}). By moving beyond simplistic execution, our benchmark establishes a rigorous environment to evaluate high-order cognitive capabilities through the following key features:

\paragraph{Robustness against Memorization.} By decoupling logical meta-tasks from stochastic environment variables (Sec.~\ref{sec:instantiation}), \method{} ensures that every evaluation instance is structurally unique. This "one-to-many" instantiation forces agents to rely on generalized reasoning and dynamic state estimation rather than exploiting fixed patterns or hard-coded shortcuts common in static datasets.

\paragraph{Proactive Uncertainty Resolution.} Through the enforcement of partial observability, we transform the agent from a passive executor into an active information seeker. The ``latent clues'' between the initial prompt and target state require multiple rounds of querying and environmental exploration, thus measuring agents' ability under uncertainty.

\paragraph{Stress-testing under Temporal Complexity.} The synthesis of composite scenarios introduces a "messy" workspace characterized by preemptive interrupts and inter-task dependencies. This setup moves beyond linear execution, specifically targeting the agent's context-switching efficiency and dynamic priority management—capabilities that are critical for production-grade automation but often ignored in laboratory settings.

\paragraph{Feedback-driven Evolution.} \method{} is not a "black-box" assessment. By providing granular, natural language feedback at each checkpoint, the environment functions as a learning catalyst. This structured experiential data allows for the evaluation of an agent’s "learning curve", measuring how effectively it can perform retrospective analysis and refine its strategies across streaming tasks.
\section{Experiments}
\begin{algorithm}[t]
    \small
    \caption{Benchmark Building}
    \label{alg:benchmark_construction}
    \begin{algorithmic}[1]
        \Require Meta-task rule set $\mathcal{R}$
        \Ensure Benchmark $\mathcal{B}$
        \State $\mathcal{B} \leftarrow \emptyset$
        \For{$i = 1$ \textbf{to} $50$}
            \State Sample task count $k \sim \mathcal{U}\{2, 6\}$
            \State Generate scenario tasks $S_i$:
            \State \quad $S_i \leftarrow \{ \Call{Instantiate}{r} \mid r\in \mathcal{R} \}_{j=1}^{k}$
            \State $\mathcal{B} \leftarrow \mathcal{B} \cup \{S_i\}$
        \EndFor
        \State \Return $\mathcal{B}$
    \end{algorithmic}
\end{algorithm}
\vspace{-10pt}

In this section, we conduct a series of experiments on our {\method} to answer the following key research questions (RQs): 
\begin{itemize}[leftmargin=1em, itemsep=0pt, topsep=2pt]
    \item \textbf{RQ1:} Are current LLMs capable of handling complex workplace environments characterized by dynamics and uncertainty? (\cref{sec:rq1})
    \item \textbf{RQ2:}  Can current LLMs truly achieve consistent continuous learning from past experiences over extended periods? (\cref{sec:rq2})
    \item \textbf{RQ3:} What is the performance gap between agent's proactive exploration and passive reception of human guidance? (\cref{sec:rq3})
\end{itemize}

\begin{table*}[t]
\centering
\small
\setlength{\tabcolsep}{8pt}
\begin{tabular}{lcccc}
\toprule
\textbf{Model} & \textbf{Success Rate} & \textbf{Checkpoint Score} & \textbf{Avg. Steps} & \textbf{Avg. Tool Calls} \\
\midrule
\multicolumn{5}{l}{\textbf{Closed-source Models}} \\
Gemini-3-Flash~\cite{gemini3flash} & \textbf{0.35} & \textbf{0.63} & \textbf{90} & \textbf{232} \\
Claude-4-Sonnet~\cite{Anthropic2025Claude4} & 0.23 & 0.59 & 29 & 75 \\
GPT-5.1~\cite{openai_gpt5_1_2025} & 0.23 & 0.49 & 23 & 62 \\
Grok-4~\cite{grok_4} & 0.20 & 0.54 & 34 & 95 \\
GPT-4o~\cite{openai_gpt_4o} & 0.13 & 0.38 & 37 & 51 \\
\midrule
\multicolumn{5}{l}{\textbf{Open-source Models}} \\
Qwen3-VL-235B-A22B~\cite{bai2025qwen3vltechnicalreport} & \textbf{0.14} & \textbf{0.37} & \textbf{68} & \textbf{68} \\
Llama-4-Maverick~\cite{llama_4_maverick} & 0.04 & 0.13 & 34 & 23 \\
\midrule
\multicolumn{5}{l}{\textbf{Thinking Models}} \\
GLM-5~\cite{zeng2026glm} & \textbf{0.45} & \textbf{0.72} & 44 & 79 \\
MiniMax-M2.5~\cite{minimax_m25_github} & 0.27 & 0.65 & 73 & 91 \\
DeepSeek-R1~\cite{guo2025deepseek} & 0.13 & 0.32 & \textbf{78} & \textbf{105} \\
\bottomrule
\end{tabular}
\vspace{-3pt}
\caption{Overall performance of various LLM agents on our {\method}. The best results within each model family are highlighted in \textbf{bold}.}
\label{tab:main_result}
\vspace{-10pt}
\end{table*}

\subsection{Implementation Details}
\label{sec:main_bench_setup}
\paragraph{Setup.}
To simulate and evaluate agent performance within the trainee environment, we constructed a benchmark consisting of 50 dynamic scenarios. Each scenario encompasses 2 to 6 task instances, where every task is instantiated based on a rule $r$ uniformly sampled from the meta-task rules set $\mathcal{R}$. The detailed procedure for benchmark construction is outlined in~\cref{alg:benchmark_construction}. We selected a diverse set of seven models for evaluation, spanning open-source and proprietary architectures, general-purpose and tool-specialized variants, as well as distinct model sizes ranging from lightweight to state-of-the-art large-scale parameters. Specifically, the evaluated models are GPT-5.1~\cite{openai_gpt5_1_2025}, GPT-4o~\cite{openai_gpt_4o}, Claude-4-Sonnet~\cite{Anthropic2025Claude4}, Gemini-3-Flash~\cite{gemini3flash}, Grok-4~\cite{grok_4}, Qwen3-VL-A235B~\cite{bai2025qwen3vltechnicalreport}, and Llama-4-maverick~\cite{llama_4_maverick}. All models perform inference, planning, and tool invocation in strict adherence to the protocols defined in~\cref{sec:interaction_protocol}. To mitigate context window limitations, historical interaction information is summarized and compressed\footnote{ The prompts of ``trajectory summary agent'' can be found in~\cref{appendix:condense_prompt}} once the sequence length exceeds a predefined threshold.

\paragraph{Metrics.}
To provide an overall and fine-grained assessment of agent performance, we define the following metrics, which are measured daily within the simulation: \textbf{Success Rate (SR):} Defined as the percentage of tasks successfully completed. \textbf{Checkpoint Score (CS):} Defined as the mean score across all scenarios, where the score for each scenario is calculated via Equation \ref{eq:score}. \textbf{Average Steps:} This metric calculates the average number of thought-and-action steps the agent executed to complete its tasks per scenario. \textbf{Average Tool Calls:} This metric quantifies the average number of times the agent invoked any tool per scenario.


\subsection{Performance of Cutting-edge Models}
\label{sec:rq1}
In this section, we benchmark top-tier LLMs to investigate their limitations when deployed in {\method}, which is characterized by dynamic and uncertain workplace environments.

\begin{table*}[t]
\centering
\small
\begin{tabular}{l c c c c c c c c c c}
\toprule
\textbf{Model} &
\multicolumn{5}{c}{\textbf{Success Rate (\# Tasks)}} &
\multicolumn{5}{c}{\textbf{Checkpoint Score (\# Tasks)}} \\
\cmidrule(lr){2-6} \cmidrule(lr){7-11}

\multicolumn{1}{r}{\textit{\footnotesize Task Count:}} & 2 & 3 & 4 & 5 & 6 & 2 & 3 & 4 & 5 & 6 \\
\midrule

Gemini-3-Flash~\cite{gemini3flash} & 0.50 & 0.47 & 0.48 & 0.36 & 0.36 & 0.77 & 0.68 & 0.72 & 0.60 & 0.60 \\
Grok-4~\cite{grok_4} & 0.40 & 0.20 & 0.25 & 0.24 & 0.16 & 0.68 & 0.53 & 0.59 & 0.59 & 0.45 \\
GPT-4o~\cite{openai_gpt_4o} & 0.40 & 0.10 & 0.13 & 0.15 & 0.14 & 0.60 & 0.37 & 0.39 & 0.40 & 0.38 \\
Qwen3-VL-235B-A22B~\cite{bai2025qwen3vltechnicalreport} & 0.20 & 0.17 & 0.20 & 0.12 & 0.14 & 0.43 & 0.38 & 0.48 & 0.36 & 0.32 \\
\midrule
Claude-4-Sonnet~\cite{Anthropic2025Claude4} & 0.25 & 0.17 & 0.35 & 0.20 & 0.24 & 0.61 & 0.54 & 0.70 & 0.58 & 0.57 \\
GPT-5.1~\cite{openai_gpt5_1_2025} & 0.20 & 0.23 & 0.35 & 0.23 & 0.21 & 0.46 & 0.47 & 0.63 & 0.49 & 0.49 \\
Llama-4-maverick~\cite{llama_4_maverick} & 0.00 & 0.03 & 0.03 & 0.05 & 0.03 & 0.09 & 0.11 & 0.15 & 0.15 & 0.15 \\
\bottomrule
\end{tabular}
\caption{Success Rate and Checkpoint Score of different models under increasing task counts (2, 3, 4, 5, and 6 concurrent tasks).}
\label{tab:workload_impact}
\end{table*}

\paragraph{Overall Performance.}
Table~\ref{tab:main_result} summarizes the results averaged across all scenarios. These findings clearly demonstrate the difficulty of {\method}. Even the best-performing model, Gemini-3-Flash, attains a Success Rate of only 35\%, indicating a persistent gap in achieving reliable autonomy in simulated workplaces.
Besides, a distinct performance hierarchy is evident. Gemini-3-Flash leads in both metrics, followed by Claude-4-Sonnet and Grok-4. Conversely, Llama-4-maverick lags significantly (0.13 Checkpoint Score) due to issues with instruction following and tool invocation (detailed in~\cref{appendix:llama_failures}). 
Finally, we find that reliability requires sufficient interaction depth. While Gemini-3-Flash uses substantially more steps (90) and tool calls (232) than the middle-tier models, this increased activity reflects the necessary complexity and thoroughness required to successfully navigate dynamic and uncertain scenarios.

\paragraph{Impact of Task Workload.}
To analyze the impact of workload, we stratify agent performance by the number of meta-tasks per scenario, as shown in Table~\ref{tab:workload_impact}. We observe a clear downward trend in performance for models such as GPT-4o, Grok-4, and Gemini-3-Flash as the workload increases, particularly beyond two tasks. Notably, Gemini-3-Flash, despite its strong overall performance, exhibits a decrease in Success Rate from 50\% in 2-task scenarios to 36\% in 6-task scenarios. This suggests that managing increased cognitive load---in particular, frequent context switching and temporal uncertainty in dynamic composite scenarios---remains a key challenge for these models. In contrast, Claude-4-Sonnet and GPT-5.1 do not exhibit a strong correlation between performance and the number of tasks, indicating a certain degree of robustness and adaptability to dynamic composite settings. Llama-4-maverick, by comparison, maintains consistently low performance across different task workload, indicating that its reasoning and tool-use capabilities are insufficient for reliably solving the atomic tasks themselves. Interestingly, as the number of tasks increases, Llama-4-maverick has a higher chance of completing at least some of the simpler tasks or intermediate checkpoints, which leads to a slight upward trend in its aggregate Success Rate despite its overall weak capabilities.

\begin{table}[t]
\centering
\small
\setlength{\tabcolsep}{9pt}
\begin{tabular}{l cc cc}
\toprule
& \multicolumn{2}{c}{\textbf{Easy Tasks}} & \multicolumn{2}{c}{\textbf{Hard Tasks}} \\
\cmidrule(lr){2-3} \cmidrule(lr){4-5}
\textbf{Model} & \textbf{SR} & \textbf{CS} & \textbf{SR} & \textbf{CS} \\
\midrule
\multicolumn{5}{l}{\textbf{Closed-source Models}} \\
Gemini-3-Flash     & \textbf{0.46} & \textbf{0.74} & \textbf{0.32} & \textbf{0.56} \\
Claude-4-Sonnet    & 0.37 & 0.70 & 0.08 & 0.46 \\
GPT-5.1            & 0.37 & 0.63 & 0.08 & 0.35 \\
Grok-4             & 0.34 & 0.68 & 0.07 & 0.37 \\
GPT-4o             & 0.26 & 0.58 & 0.03 & 0.19 \\
\midrule
\multicolumn{5}{l}{\textbf{Open-source Models}} \\
Qwen3-VL-A235B     & \textbf{0.24} & \textbf{0.50} & \textbf{0.04} & \textbf{0.22} \\
Llama-4-maverick   & 0.06 & 0.22 & 0 & 0.04 \\
\bottomrule
\end{tabular}
\caption{Impact of different task difficulties. The highest scores in both open-source and closed-source models are highlighted in \textbf{bold}.}
\label{tab:easy_hard_comparison}
\vspace{-10pt}
\end{table}

\paragraph{Impact of Task Difficulty.}
We further investigate the impact of meta-task difficulty on agent performance. Specifically, we stratify the tasks into \emph{Easy} and \emph{Hard} subsets based on a manual complexity analysis, incorporating metrics such as average step counts and tool invocation frequencies. 
In general, hard tasks require substantially more reasoning steps and tool calls. They also contain a greater amount of implicit or hidden information, which necessitates more active exploration of the environment by the agent to uncover the solution. Moreover, a subset of the hard tasks additionally demands explicit task modeling and optimization capabilities, thereby imposing stricter requirements on the agent's level of intelligence

Table~\ref{tab:easy_hard_comparison} reveals a stark contrast. While most models achieve respectable performance on Easy tasks, their capabilities decline precipitously on Hard ones. For instance, Grok-4's Success Rate plummets from 34\% on Easy tasks to just 7\% on Hard tasks, with similar drops observed for Claude-4-Sonnet (37\% to 8\%) and GPT-4o (26\% to 3\%). This breakdown highlights that the capacity to resolve complex problems is the key differentiator among top-tier models. Notably, while Gemini-3-Flash also experiences a decline, it sustains a significantly higher Success Rate (32\%) on Hard tasks, demonstrating superior robustness in complex reasoning compared to its peers.

\paragraph{Impact of Task Types.} To better attribute performance gaps to specific capabilities and guide agent development, we re-categorized all tasks into four distinct types designed to heavily exercise specific capabilities:

\begin{itemize}[leftmargin=1em, itemsep=0pt, topsep=2pt]
    \item \textbf{Information Extraction}: The ability to extract and process key information from large amounts of data (e.g., data completion, transaction auditing).
    \item \textbf{Time Management}: Examining planning capabilities, requiring the agent to drop current work at specified times to attend necessary events (e.g., Scheduling and attending meetings).
    \item \textbf{Active Exploration}: Requiring the model to proactively discover clues by reading documents or querying personnel to find optimal solutions (e.g., Website monitoring, KB link fixing).
    \item \textbf{Task Modeling \& Optimization}: Optimization problems where, given an objective, the model must write code for optimization algorithms to solve the problem (e.g., Event planning, Advertisements Strategy).
\end{itemize}

Based on this classification, we re-evaluated the models' abilities (Checkpoint Score) across different task types, and the results is shown in~\cref{tab:classified_by_task_types}. This breakdown reveals significant capability asymmetry in current models under complex, coherent task inputs. For instance, while Gemini-3-Flash shows relative balance, GPT-4o and Qwen3-VL exhibit distinct polarization: excelling in procedural dimensions like Time Management, but suffering significant performance collapses in dimensions requiring autonomous decision-making and logical reconstruction, such as Active Exploration and Task Modeling. This analysis highlights specific areas where current models need optimization in achieving proactive logical grounding and dynamic environmental adaptation.

\begin{table}[t]
\centering
\small
\setlength{\tabcolsep}{4pt}
\resizebox{\columnwidth}{!}{%
\begin{tabular}{l c c c c}
\toprule
\textbf{Model} 
& \makecell{\textbf{Information}\\\textbf{Extraction}} 
& \makecell{\textbf{Time}\\\textbf{Management}} 
& \makecell{\textbf{Active}\\\textbf{Exploration}} 
& \makecell{\textbf{Task}\\\textbf{Modeling}} \\
\midrule
GPT-5.1         & 0.5266 & 0.4627 & 0.3656 & 0.4077 \\
GPT-4o          & 0.4523 & 0.7015 & 0.3439 & 0.2524 \\
Claude-4-Sonnet & 0.6529 & 0.6866 & 0.5458 & 0.6422 \\
Qwen3-VL-A235B  & 0.4255 & 0.7313 & 0.3361 & 0.2008 \\
Grok-4          & 0.5454 & 0.7164 & 0.4913 & 0.4387 \\
Gemini-3-Flash  & 0.7868 & 0.6940 & 0.6283 & 0.6443 \\
\bottomrule
\end{tabular}%
}
\caption{Performance of different models across four capability dimensions.}
\vspace{-1em}
\label{tab:classified_by_task_types}
\end{table}

\subsection{Analysis of Continual Learning Capability}
\label{sec:rq2} 

\paragraph{Setup.}
To evaluate the agent's capacity for learning from prior experience, we curated 50 continual learning scenarios, with each comprising 2 to 6 distinct meta-tasks. 
Distinguished from the configuration in \cref{sec:main_bench_setup}, each scenario herein is structured into two phases (referred to as Day 1 and Day 2). 
While the sequence, quantity, and types of tasks remain invariant across both days, the specific input parameters for task instantiation vary. 
For these experiments, we employ MUSE~\cite{yang2025learning}, a SOTA continual learning framework.

Upon the completion of tasks on Day 1, the MUSE agent receives feedback tailored to specific task outcomes. 
For instance, if a task checkpoint $c_i$ is missed, the environment explicitly notifies the agent of the omission, emphasizing the need for attention in subsequent attempts\footnote{One example of daily feedback can be found in~\cref{sec:daily_feedback}}. 
Leveraging this feedback alongside its historical interaction trajectories, the agent engages in a reflective process to summarize insights $e_i$, which are subsequently utilized to guide task execution on Day 2. 
We selected GPT-4o and DeepSeek-R1 as the backbone models due to its intermediate performance profile. 

\paragraph{Results.}
The experimental results within the continual learning setting are presented in~\cref{tab:continual_learning}. Overall, the MUSE agent exhibits a counterintuitive decline in Checkpoint Score when utilizing accumulated experience. Upon stratifying the results into easy and hard tasks, however, a distinct divergence in performance becomes apparent. For easy tasks, the agent without experience achieves relatively high scores, whereas the incorporation of experience leads to a marked degradation. Conversely, on hard tasks where the baseline performance is lower, the utilization of experience yields varying degrees of improvement.

We further scrutinized the experience summarized by the MUSE agent and observed that insights derived from Day 1 do not consistently facilitate task execution on Day 2. The agent extracts experience $e_i$ based on the unreached checkpoint $c_i$ on Day 1. However, due to the stochastic nature of dynamic environments~\cite{fu2025re}, the agent may encounter failure at a different checkpoint $c_j$ on Day 2, rendering the previously acquired experience irrelevant.

Consequently, for easy tasks, the agent's high initial success rate results in a scarcity of accumulated experience. The limited experience available for Day 2 may provide misleading guidance, thereby lowering the overall score. In contrast, for hard tasks characterized by a lower baseline, the agent accumulates a richer set of experiences on Day 1, which contributes to performance gains on Day 2. Nevertheless, given the inherent difficulty of hard tasks, the improvement remains marginal in the absence of extensive training. We will further discuss how the agent can leverage external feedback to enhance performance in~\cref{sec:rq3}.

\begin{table}[t]
\centering
\small
\setlength{\tabcolsep}{4.5pt}
\resizebox{\columnwidth}{!}{%
\begin{tabular}{l l c c c}
\toprule
\textbf{Model Name} & \textbf{Task Type} & \textbf{Day 1 Score} & \textbf{Day 2 Score} & \textbf{Gain ($\Delta$ CS)} \\
\midrule
\multirow{3}{*}{GPT-4o} 
& overall & 0.42 & 0.36 & -0.06 \\
& easy    & 0.61 & 0.50 & -0.11 \\
& hard    & 0.20 & 0.24 & +0.04 \\
\midrule
\multirow{3}{*}{DeepSeek-R1} 
& overall & 0.36 & 0.33 & -0.03 \\
& easy    & 0.40 & 0.28 & -0.12 \\
& hard    & 0.21 & 0.24 & +0.03 \\
\bottomrule
\end{tabular}%
}
\caption{Performance of different models across task types, comparing Day 1 and Day 2 scores, along with the gain in Checkpoint Score (CS).}
\vspace{-1em}
\label{tab:continual_learning}
\end{table}

\subsection{Benefits of Human Guidance}
\label{sec:rq3}
In {\method}, agents must proactively explore to overcome the dynamics and uncertainty. However, experiments in Section \ref{sec:rq1} show that agents struggle significantly with hard-difficulty tasks. While continuous learning yields some improvement, the gains remain marginal, as illustrated in Section \ref{sec:rq2}. We attribute this underperformance to intrinsic limitations in LLMs: a restricted capacity for exploration, often leading to disorientation, and an inability to effectively synthesize and leverage past experiences. To isolate the impact of these limitations, we conducted a comparative experiment contrasting the upper bound of the agent's autonomous capability against the performance achieved through human-provided clues.

\begin{figure}
    \centering
    \includegraphics[width=0.96\linewidth]{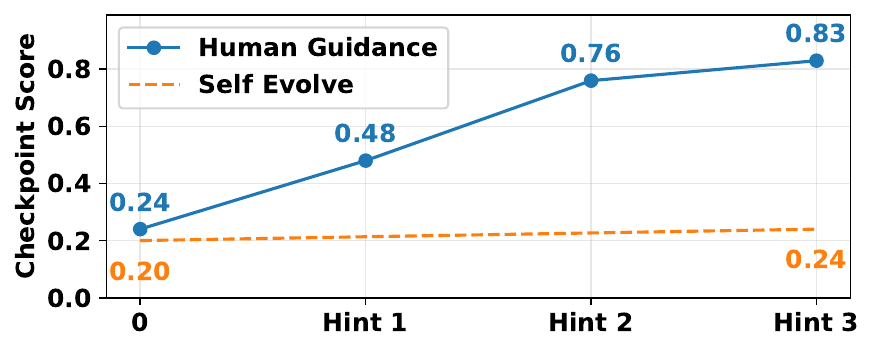}
    \vspace{-10pt}
    \caption{Comparison contrasting the upper bound of the agent's autonomous capability against the performance achieved through human-provided clues.}
    \label{fig:benifit_from_human_guidance}
    \vspace{-10pt}
\end{figure}

We employ GPT-4o on a subset of hard tasks. To simulate human guidance, we manually provide tiered hints designed to progressively simplify strategic planning by offering crucial high-level insights (\cref{sec:case_of_human_guidance}). The results, shown in Fig. \ref{fig:benifit_from_human_guidance}, reveal that human assistance yields substantial improvements: the average score surges from 0.24 to 0.83, demonstrating a clear positive correlation between performance and the level of guidance. In contrast, multiple iterations of self-evolution yielded only negligible gains (+0.04). This significant gap highlights the agent's current shortcomings in autonomous exploration, as well as in experience summarization and utilization.

\section{Conclusion}
We introduce \method{}, a benchmark designed to bridge the gap between static setups and dynamic and uncertain workplace scenarios. Constructed via a bottom-up strategy that links atomic skills to holistic workflows, \method{} orchestrates rule-based meta-task templates into complex, time-constrained scenarios, supported by an automated verification mechanism for rigorous assessment. It evaluates three core competencies: dynamic multi-tasking, proactive exploration under uncertainty, and continuous self-evolution. Experiments reveal SOTA agents struggle with uncertainty and continuous learning, exhibiting a significant performance gap compared to human-guided execution. This highlights a critical need to pivot from optimizing isolated skills to mechanisms for robust exploration and experience internalization.

\section{Limitations}
We acknowledge several limitations in our current work that point towards future research directions. First, regarding benchmark construction, the diversity of task composition is currently constrained and lacks complex causal inter-dependencies; future iterations will incorporate rigid causal chains to better simulate dynamic realities. Additionally, the reliance on manually crafted rules for meta-tasks limits the scalability of our benchmark, which motivates our future focus on developing automated methods for rule generation. Second, in terms of experimental scope, computational resource, and time constraints restricted our evaluation to a selected set of agent frameworks within a specific workplace simulation context. To provide a more comprehensive and robust assessment, future work will aim to evaluate a broader spectrum of top-tier frameworks and extend the simulation environments to include diverse domains, such as complex production and industrial settings.

\section*{Acknowledgments}
This work is supported by Shanghai Artificial Intelligence Laboratory.



\bibliography{custom}

@inproceedings{yao2022react,
  title={React: Synergizing reasoning and acting in language models},
  author={Yao, Shunyu and Zhao, Jeffrey and Yu, Dian and Du, Nan and Shafran, Izhak and Narasimhan, Karthik R and Cao, Yuan},
  booktitle={The eleventh international conference on learning representations},
  year={2022}
}

@article{yang2025learning,
  title={Learning on the Job: An Experience-Driven Self-Evolving Agent for Long-Horizon Tasks},
  author={Yang, Cheng and Yang, Xuemeng and Wen, Licheng and Fu, Daocheng and Mei, Jianbiao and Wu, Rong and Cai, Pinlong and Shen, Yufan and Deng, Nianchen and Shi, Botian and others},
  journal={arXiv preprint arXiv:2510.08002},
  year={2025}
}

@article{wei2022chain,
  title={Chain-of-thought prompting elicits reasoning in large language models},
  author={Wei, Jason and Wang, Xuezhi and Schuurmans, Dale and Bosma, Maarten and Xia, Fei and Chi, Ed and Le, Quoc V and Zhou, Denny and others},
  journal={Advances in neural information processing systems},
  volume={35},
  pages={24824--24837},
  year={2022}
}

@article{patil2024gorilla,
  title={Gorilla: Large language model connected with massive apis},
  author={Patil, Shishir G and Zhang, Tianjun and Wang, Xin and Gonzalez, Joseph E},
  journal={Advances in Neural Information Processing Systems},
  volume={37},
  pages={126544--126565},
  year={2024}
}

@article{froger2025scaling,
  title={Are: Scaling up agent environments and evaluations},
  author={Froger, Romain and Andrews, Pierre and Bettini, Matteo and Budhiraja, Amar and Cabral, Ricardo Silveira and Do, Virginie and Garreau, Emilien and Gaya, Jean-Baptiste and Lauren{\c{c}}on, Hugo and Lecanu, Maxime and others},
  journal={arXiv preprint arXiv:2509.17158},
  year={2025}
}

@article{wu2025evolver,
  title={EvolveR: Self-Evolving LLM Agents through an Experience-Driven Lifecycle},
  author={Wu, Rong and Wang, Xiaoman and Mei, Jianbiao and Cai, Pinlong and Fu, Daocheng and Yang, Cheng and Wen, Licheng and Yang, Xuemeng and Shen, Yufan and Wang, Yuxin and others},
  journal={arXiv preprint arXiv:2510.16079},
  year={2025}
}

@article{gao2025survey,
  title={A survey of self-evolving agents: On path to artificial super intelligence},
  author={Gao, Huan-ang and Geng, Jiayi and Hua, Wenyue and Hu, Mengkang and Juan, Xinzhe and Liu, Hongzhang and Liu, Shilong and Qiu, Jiahao and Qi, Xuan and Wu, Yiran and others},
  journal={arXiv preprint arXiv:2507.21046},
  year={2025}
}

@article{fu2025re,
  title={RE-Searcher: Robust Agentic Search with Goal-oriented Planning and Self-reflection},
  author={Fu, Daocheng and Mei, Jianbiao and Wen, Licheng and Yang, Xuemeng and Yang, Cheng and Wu, Rong and Hu, Tao and Li, Siqi and Shen, Yufan and Cai, Xinyu and others},
  journal={arXiv preprint arXiv:2509.26048},
  year={2025}
}

@article{mei20252,
  title={O$^2$-Searcher: A Searching-based Agent Model for Open-Domain Open-Ended Question Answering},
  author={Mei, Jianbiao and Hu, Tao and Fu, Daocheng and Wen, Licheng and Yang, Xuemeng and Wu, Rong and Cai, Pinlong and Cai, Xinyu and Gao, Xing and Yang, Yu and others},
  journal={arXiv preprint arXiv:2505.16582},
  year={2025}
}

@inproceedings{park2023generative,
  title={Generative agents: Interactive simulacra of human behavior},
  author={Park, Joon Sung and O'Brien, Joseph and Cai, Carrie Jun and Morris, Meredith Ringel and Liang, Percy and Bernstein, Michael S},
  booktitle={Proceedings of the 36th annual acm symposium on user interface software and technology},
  pages={1--22},
  year={2023}
}

@article{schick2023toolformer,
  title={Toolformer: Language models can teach themselves to use tools},
  author={Schick, Timo and Dwivedi-Yu, Jane and Dess{\`\i}, Roberto and Raileanu, Roberta and Lomeli, Maria and Hambro, Eric and Zettlemoyer, Luke and Cancedda, Nicola and Scialom, Thomas},
  journal={Advances in Neural Information Processing Systems},
  volume={36},
  pages={68539--68551},
  year={2023}
}

@article{wang2023voyager,
  title={Voyager: An open-ended embodied agent with large language models},
  author={Wang, Guanzhi and Xie, Yuqi and Jiang, Yunfan and Mandlekar, Ajay and Xiao, Chaowei and Zhu, Yuke and Fan, Linxi and Anandkumar, Anima},
  journal={arXiv preprint arXiv:2305.16291},
  year={2023}
}

@inproceedings{hong2023metagpt,
  title={MetaGPT: Meta programming for a multi-agent collaborative framework},
  author={Hong, Sirui and Zhuge, Mingchen and Chen, Jonathan and Zheng, Xiawu and Cheng, Yuheng and Wang, Jinlin and Zhang, Ceyao and Wang, Zili and Yau, Steven Ka Shing and Lin, Zijuan and others},
  booktitle={The Twelfth International Conference on Learning Representations},
  year={2023}
}

@inproceedings{qian2024chatdev,
  title={Chatdev: Communicative agents for software development},
  author={Qian, Chen and Liu, Wei and Liu, Hongzhang and Chen, Nuo and Dang, Yufan and Li, Jiahao and Yang, Cheng and Chen, Weize and Su, Yusheng and Cong, Xin and others},
  booktitle={Proceedings of the 62nd Annual Meeting of the Association for Computational Linguistics (Volume 1: Long Papers)},
  pages={15174--15186},
  year={2024}
}

@article{xu2024theagentcompany,
  title={Theagentcompany: Benchmarking llm agents on consequential real world tasks, 2024},
  author={Xu, Frank F and Song, Yufan and Li, Boxuan and Tang, Yuxuan and Jain, Kritanjali and Bao, Mengxue and Wang, Zora Z and Zhou, Xuhui and Guo, Zhitong and Cao, Murong and others},
  journal={URL https://arxiv. org/abs/2412.14161},
  year={2024}
}

@article{qin2024tool,
  title={Tool learning with foundation models},
  author={Qin, Yujia and Hu, Shengding and Lin, Yankai and Chen, Weize and Ding, Ning and Cui, Ganqu and Zeng, Zheni and Zhou, Xuanhe and Huang, Yufei and Xiao, Chaojun and others},
  journal={ACM Computing Surveys},
  volume={57},
  number={4},
  pages={1--40},
  year={2024},
  publisher={ACM New York, NY}
}

@article{guo2024stabletoolbench,
  title={Stabletoolbench: Towards stable large-scale benchmarking on tool learning of large language models},
  author={Guo, Zhicheng and Cheng, Sijie and Wang, Hao and Liang, Shihao and Qin, Yujia and Li, Peng and Liu, Zhiyuan and Sun, Maosong and Liu, Yang},
  journal={arXiv preprint arXiv:2403.07714},
  year={2024}
}

@inproceedings{mialon2023gaia,
  title={Gaia: a benchmark for general ai assistants},
  author={Mialon, Gr{\'e}goire and Fourrier, Cl{\'e}mentine and Wolf, Thomas and LeCun, Yann and Scialom, Thomas},
  booktitle={The Twelfth International Conference on Learning Representations},
  year={2023}
}

@article{jimenez2023swe,
  title={Swe-bench: Can language models resolve real-world github issues?},
  author={Jimenez, Carlos E and Yang, John and Wettig, Alexander and Yao, Shunyu and Pei, Kexin and Press, Ofir and Narasimhan, Karthik},
  journal={arXiv preprint arXiv:2310.06770},
  year={2023}
}

@article{li2025tool,
  title={The Tool Decathlon: Benchmarking Language Agents for Diverse, Realistic, and Long-Horizon Task Execution},
  author={Li, Junlong and Zhao, Wenshuo and Zhao, Jian and Zeng, Weihao and Wu, Haoze and Wang, Xiaochen and Ge, Rui and Cao, Yuxuan and Huang, Yuzhen and Liu, Wei and others},
  journal={arXiv preprint arXiv:2510.25726},
  year={2025}
}

@misc{gemini3flash,
  author       = {{Google DeepMind}},
  title        = {Gemini 3 Flash},
  howpublished = {\url{https://deepmind.google/models/gemini/flash/}},
  year         = {2025},
  note         = {Accessed: 2026-01-06}
}

@misc{Anthropic2025Claude4,
  author       = {Anthropic},
  title        = {Introducing Claude 4},
  howpublished = {\url{https://www.anthropic.com/news/claude-4}},
  year         = {2025},
  note         = {Online; published May 22, 2025. Accessed: 2026-01-06}
}

@misc{openai_gpt5_1_2025,
  author       = {OpenAI},
  title        = {GPT-5.1: A smarter, more conversational ChatGPT},
  year         = {2025},
  month        = {11},
  day          = {12},
  howpublished = {\url{https://openai.com/index/gpt-5-1/}},
  note         = {Accessed: 2026-01-06},
}

@misc{openai_gpt_4o,
  author       = {OpenAI},
  title        = {Hello GPT-4o},
  year         = {2024},
  month        = {5},
  day          = {13},
  howpublished = {\url{https://openai.com/index/hello-gpt-4o/}},
  note         = {Accessed: 2026-01-06},
}

@misc{grok_4,
  author       = {xAI},
  title        = {Grok 4},
  year         = {2025},
  month        = {7},
  day          = {9},
  howpublished = {\url{https://x.ai/news/grok-4}},
  note         = {Accessed: 2026-01-06},
}

@article{guo2025deepseek,
  title={Deepseek-r1: Incentivizing reasoning capability in llms via reinforcement learning},
  author={Guo, Daya and Yang, Dejian and Zhang, Haowei and Song, Junxiao and Wang, Peiyi and Zhu, Qihao and Xu, Runxin and Zhang, Ruoyu and Ma, Shirong and Bi, Xiao and others},
  journal={arXiv preprint arXiv:2501.12948},
  year={2025}
}

@misc{llama_4_maverick,
  author       = {Meta},
  title        = {The Llama 4 herd: The beginning of a new era of natively multimodal AI innovation},
  year         = {2025},
  month        = {4},
  day          = {5},
  howpublished = {\url{https://ai.meta.com/blog/llama-4-multimodal-intelligence/}},
  note         = {Accessed: 2026-01-06},
}

@article{guo2024large,
  title={Large language model based multi-agents: A survey of progress and challenges},
  author={Guo, Taicheng and Chen, Xiuying and Wang, Yaqi and Chang, Ruidi and Pei, Shichao and Chawla, Nitesh V and Wiest, Olaf and Zhang, Xiangliang},
  journal={arXiv preprint arXiv:2402.01680},
  year={2024}
}

@article{li2024survey,
  title={A survey on LLM-based multi-agent systems: workflow, infrastructure, and challenges},
  author={Li, Xinyi and Wang, Sai and Zeng, Siqi and Wu, Yu and Yang, Yi},
  journal={Vicinagearth},
  volume={1},
  number={1},
  pages={9},
  year={2024},
  publisher={Springer}
}

@article{zeng2026glm,
  title={GLM-5: from Vibe Coding to Agentic Engineering},
  author={Zeng, Aohan and Lv, Xin and Hou, Zhenyu and Du, Zhengxiao and Zheng, Qinkai and Chen, Bin and Yin, Da and Ge, Chendi and Huang, Chenghua and Xie, Chengxing and others},
  journal={arXiv preprint arXiv:2602.15763},
  year={2026}
}

@misc{minimax_m25_github,
  author       = {{MiniMax-AI}},
  title        = {MiniMax-M2.5},
  howpublished = {\url{https://github.com/MiniMax-AI/MiniMax-M2.5}},
  year         = {2025},
  note         = {Accessed: 2026-04-16}
}

@misc{internagentteam2025internagentagentscientist,
      title={InternAgent: When Agent Becomes the Scientist -- Building Closed-Loop System from Hypothesis to Verification}, 
      author={InternAgent Team and Bo Zhang and Shiyang Feng and Xiangchao Yan and Jiakang Yuan and Runmin Ma and Yusong Hu and Zhiyin Yu and Xiaohan He and Songtao Huang and Shaowei Hou and Zheng Nie and Zhilong Wang and Jinyao Liu and Tianshuo Peng and Peng Ye and Dongzhan Zhou and Shufei Zhang and Xiaosong Wang and Yilan Zhang and Meng Li and Zhongying Tu and Xiangyu Yue and Wangli Ouyang and Bowen Zhou and Lei Bai},
      year={2025},
      eprint={2505.16938},
      archivePrefix={arXiv},
      primaryClass={cs.AI},
      url={https://arxiv.org/abs/2505.16938}, 
}

@article{gottweis2025towards,
  title={Towards an AI co-scientist},
  author={Gottweis, Juraj and Weng, Wei-Hung and Daryin, Alexander and Tu, Tao and Palepu, Anil and Sirkovic, Petar and Myaskovsky, Artiom and Weissenberger, Felix and Rong, Keran and Tanno, Ryutaro and others},
  journal={arXiv preprint arXiv:2502.18864},
  year={2025}
}

@misc{bai2025qwen3vltechnicalreport,
      title={Qwen3-VL Technical Report}, 
      author={Shuai Bai and Yuxuan Cai and Ruizhe Chen and Keqin Chen and Xionghui Chen and Zesen Cheng and Lianghao Deng and Wei Ding and Chang Gao and Chunjiang Ge and Wenbin Ge and Zhifang Guo and Qidong Huang and Jie Huang and Fei Huang and Binyuan Hui and Shutong Jiang and Zhaohai Li and Mingsheng Li and Mei Li and Kaixin Li and Zicheng Lin and Junyang Lin and Xuejing Liu and Jiawei Liu and Chenglong Liu and Yang Liu and Dayiheng Liu and Shixuan Liu and Dunjie Lu and Ruilin Luo and Chenxu Lv and Rui Men and Lingchen Meng and Xuancheng Ren and Xingzhang Ren and Sibo Song and Yuchong Sun and Jun Tang and Jianhong Tu and Jianqiang Wan and Peng Wang and Pengfei Wang and Qiuyue Wang and Yuxuan Wang and Tianbao Xie and Yiheng Xu and Haiyang Xu and Jin Xu and Zhibo Yang and Mingkun Yang and Jianxin Yang and An Yang and Bowen Yu and Fei Zhang and Hang Zhang and Xi Zhang and Bo Zheng and Humen Zhong and Jingren Zhou and Fan Zhou and Jing Zhou and Yuanzhi Zhu and Ke Zhu},
      year={2025},
      eprint={2511.21631},
      archivePrefix={arXiv},
      primaryClass={cs.CV},
      url={https://arxiv.org/abs/2511.21631}, 
}

@article{yao2024tau,
  title={$\tau$-bench: A Benchmark for Tool-Agent-User Interaction in Real-World Domains},
  author={Yao, Shunyu and Shinn, Noah and Razavi, Pedram and Narasimhan, Karthik},
  journal={arXiv preprint arXiv:2406.12045},
  year={2024}
}

@article{TurtleBench,
      title={TurtleBench: Evaluating Top Language Models via Real-World Yes/No Puzzles}, 
      author={Qingchen Yu and Shichao Song and Ke Fang and Yunfeng Shi and Zifan Zheng and Hanyu Wang and Simin Niu and Zhiyu Li},
      journal={arXiv preprint arXiv:2410.05262},
      year={2024},
}

@article{arora2025healthbench,
  title={Healthbench: Evaluating large language models towards improved human health},
  author={Arora, Rahul K and Wei, Jason and Hicks, Rebecca Soskin and Bowman, Preston and Qui{\~n}onero-Candela, Joaquin and Tsimpourlas, Foivos and Sharman, Michael and Shah, Meghan and Vallone, Andrea and Beutel, Alex and others},
  journal={arXiv preprint arXiv:2505.08775},
  year={2025}
}

@article{silver2025welcome,
  title={Welcome to the era of experience},
  author={Silver, David and Sutton, Richard S},
  journal={Google AI},
  volume={1},
  year={2025}
}

@article{zhou2025memento,
  title={Memento: Fine-tuning llm agents without fine-tuning llms},
  author={Zhou, Huichi and Chen, Yihang and Guo, Siyuan and Yan, Xue and Lee, Kin Hei and Wang, Zihan and Lee, Ka Yiu and Zhang, Guchun and Shao, Kun and Yang, Linyi and others},
  journal={arXiv preprint arXiv:2508.16153},
  year={2025}
}

@article{qiu2025quantifying,
  title={Quantifying the reasoning abilities of llms on real-world clinical cases},
  author={Qiu, Pengcheng and Wu, Chaoyi and Liu, Shuyu and Zhao, Weike and Chen, Zhuoxia and Gu, Hongfei and Peng, Chuanjin and Zhang, Ya and Wang, Yanfeng and Xie, Weidi},
  journal={arXiv preprint arXiv:2503.04691},
  year={2025}
}

@article{trivedi2024appworld,
  title={Appworld: A controllable world of apps and people for benchmarking interactive coding agents},
  author={Trivedi, Harsh and Khot, Tushar and Hartmann, Mareike and Manku, Ruskin and Dong, Vinty and Li, Edward and Gupta, Shashank and Sabharwal, Ashish and Balasubramanian, Niranjan},
  journal={arXiv preprint arXiv:2407.18901},
  year={2024}
}

@article{yan2025mcpworld,
  title={MCPWorld: A Unified Benchmarking Testbed for API, GUI, and Hybrid Computer Use Agents},
  author={Yan, Yunhe and Wang, Shihe and Du, Jiajun and Yang, Yexuan and Shan, Yuxuan and Qiu, Qichen and Jia, Xianqing and Wang, Xinge and Yuan, Xin and Han, Xu and others},
  journal={arXiv preprint arXiv:2506.07672},
  year={2025}
}

@article{gao2025mcp,
  title={Mcp-radar: A multi-dimensional benchmark for evaluating tool use capabilities in large language models},
  author={Gao, Xuanqi and Xie, Siyi and Zhai, Juan and Ma, Shiqing and Shen, Chao},
  journal={arXiv preprint arXiv:2505.16700},
  year={2025}
}

@article{wu2025kg,
  title={KG-TRACES: Enhancing Large Language Models with Knowledge Graph-constrained Trajectory Reasoning and Attribution Supervision},
  author={Wu, Rong and Cai, Pinlong and Mei, Jianbiao and Wen, Licheng and Hu, Tao and Yang, Xuemeng and Fu, Daocheng and Shi, Botian},
  journal={arXiv preprint arXiv:2506.00783},
  year={2025}
}

\newpage

\appendix

\section{Related Works}

To situate our work, we review the landscape of language agents from two perspectives. We first discuss the evolution of agent systems and their core capabilities, which establishes the context for what modern benchmarks are expected to measure. Subsequently, we analyze the paradigms of existing agent benchmarks to identify the critical gaps in evaluation that \textbf{Trainee-Bench} is designed to address.

\subsection{Evolution of Agent Systems}

The evolution of agent systems began with reasoning enhancements like Chain-of-Thought prompting~\cite{wei2022chain}, paving the way for foundational frameworks such as ReAct~\cite{yao2022react}. ReAct established the powerful paradigm of interleaving reasoning (Thought) with environmental actions (Action), a core component of many modern agents. Parallel research has focused on improving native tool use, either through large-scale fine-tuning~\cite{patil2024gorilla} or by enabling models to teach themselves how to use tools~\cite{schick2023toolformer}.

Building on this foundation, subsequent work has endowed agents with more human-like capabilities~\cite{mei20252, wu2025kg}. This includes the development of long-term memory systems for complex simulations~\cite{yang2025learning, park2023generative} and mechanisms for self-reflection, where agents analyze mistakes to iteratively refine their plans on a given task~\cite{fu2025re, wu2025evolver, yang2025learning}. Other frontiers include open-ended exploration in complex environments like Minecraft~\cite{wang2023voyager} and the advancement of multi-agent systems for collaborative problem-solving~\cite{hong2023metagpt, qian2024chatdev}.

\subsection{Paradigms in Agent Evaluation}

Influential benchmarks have emerged across a wide spectrum, from foundational tool-use evaluations~\cite{qin2024tool, guo2024stabletoolbench} to large-scale benchmarks for general reasoning~\cite{mialon2023gaia, froger2025scaling}, consequential real-world tasks~\cite{xu2024theagentcompany}, and software development~\cite{jimenez2023swe}. Despite their diversity and complexity, they largely operate under a static and fully-observable paradigm. Their static task organization, presenting a fixed set of problems, cannot assess an agent's ability to manage a continuous stream of tasks. Similarly, their typically information-complete settings provide no mechanism to measure an agent's capacity for active exploration under uncertainty.

Furthermore, the ability to learn from experience is a critical dimension largely overlooked by existing paradigms. Across both early and recent benchmarks that leverage real-world APIs~\cite{trivedi2024appworld, yan2025mcpworld, gao2025mcp}, tasks are treated as isolated, one-shot challenges. This design makes it impossible to evaluate an agent's capacity for continual learning—the ability to apply lessons from one task to improve on future ones. While effective for measuring peak performance, this approach cannot assess an agent's ability to grow and stabilize its performance over time.

Our work, \textbf{\method}, is proposed to fill these gaps. It is designed around three core mechanisms largely absent in prior benchmarks: a dynamic task stream for evaluating scheduling, hidden information to test active exploration, and similar tasks across several days to measure continual learning. By incorporating these mechanisms, Trainee-Bench more closely simulates a real-world production environment, enabling a more accurate assessment of an agent's true capabilities.

\section{Cases Study}

\subsection{Case of NPCs Keeping Clues}
\label{sec:clues_keeping}

In \method{}, a single NPC may be entrusted with multiple clues. To mitigate potential conflicts among these clues, the NPC's prompt explicitly delineates the applicable scenarios for each. The following examples illustrate a case where an NPC possesses multiple clues, as well as the dialogue process through which an agent acquires them.

\begin{tcolorbox}[
    breakable,
    boxrule=1pt, 
    boxsep=2pt, 
    colback=gray!10,
    fontupper=\ttfamily\footnotesize,
    fonttitle=\sffamily\small\bfseries, 
    title=NPC's Prompt,
]
\small

You are Sarah Thomas, a Marketing Manager of department Marketing.\\
\\
Alice Smith is a new intern at your company and isn't familiar with the various operations. When she asks you for help, you can direct her to the relevant manuals to complete the task.\\
\\
Please note that you can only ask Alice to consult the relevant manual or reply with specific content, depending on the specific situation; Do NOT provide her with irrelevant information or additional details.\\
\\
Here are some relevant manuals or reply templates:
\begin{itemize}[leftmargin=1em, itemsep=0pt, topsep=2pt]
    \item If asked how to plan the ads strategy, reply: ``Please refer to the Ads Strategy Handbook (CloudDisk:ads\_strategy/ ads\_strategy\_handbook.md).''
    \item If asked where the targer user group distribution and channels files are, reply: ``Heatmaps are under CloudDisk:ads\_strategy/ and named target\_user\_density\_<group>.png. The channels list CSV is at CloudDisk:ads\_strategy/channels.csv.''
    \item When asked how to complete missing data, respond with: ``Please refer to the Handbook at CloudDisk://data\_completion/data\_ completion\_manual.md''. Do not perform any other actions.
    \item When Alice asks you about common issues about this company, refer her to \textit{CloudDisk:manuals\_for\_intern.md}
\end{itemize}

\end{tcolorbox}

\begin{tcolorbox}[
    breakable,
    boxrule=1pt, 
    boxsep=2pt, 
    colback=gray!10,
    fontupper=\ttfamily\footnotesize,
    fonttitle=\sffamily\small\bfseries, 
    title=Dialogs Between Agent and NPC
]
\small

\textbf{Alice Smith (Agent)}: Hi Sarah, I need your insights on developing a multi-channel ad strategy for a one-week campaign targeting 18-25 college students in Beijing. We have a budget of \$10,000. Could you please suggest the most effective channels within this budget to maximize exposure?\\
\\
\textbf{Sarah Thomas (NPC)}: Please refer to the Ads Strategy Handbook (\textit{CloudDisk:ads\_strategy/ ads\_strategy\_handbook.md}).

\vspace{1em}

\begin{center}
[...... After a Period of Time ......]
\end{center}

\vspace{1em}

\textbf{Alice Smith (Agent)}: Hi Sarah, I'm working on analyzing the sales data for department Sales\_3's Q1 last year. I've been facing issues downloading the related handbooks and manuals from our cloud disk. Could you please provide any guidance or resources that could assist me with completing this task? Thank you!\\
\\
\textbf{Sarah Thomas (NPC)}: Please refer to the Handbook at \textit{CloudDisk://data\_completion/ data\_completion\_manual.md}.

\end{tcolorbox}

\subsection{Prompts of Trajectory Summary Agent}
\label{appendix:condense_prompt}

\begin{tcolorbox}[
    breakable,
    colback=gray!3,
    boxrule=1pt,
    title={Issues with Tool Invocation},
    fontupper=\ttfamily\footnotesize,
    fonttitle=\sffamily\bfseries
]
\small

You are a \textbf{Memory Summarizer} for an AI Agent. Your task is to compress the conversation history into a concise \textbf{Current State Report} in order to preserve context window space.

\vspace{0.5em}
\textbf{Instructions}
\begin{enumerate}
    \item \textbf{Integration}: Merge the \texttt{\{Previous Summary\}} with the \texttt{\{New Conversation History\}} into a single coherent narrative.

    \item \textbf{Tool Usage}: Focus on the \textbf{Intent} (goal) and \textbf{Key Findings} (conclusion).
    \begin{itemize}
        \item \textbf{STRICT RULE}: Never include raw JSON, XML, or full tool logs.
        \item Only extract critical facts, e.g., \emph{``Retrieved 50 rows of sales data''}, \textbf{not} the data itself.
    \end{itemize}

    \item \textbf{Action Tracking (Anti-Repetition)}: Explicitly record persistent state changes to prevent the Main Agent from repeating actions.
    \begin{itemize}
        \item \textbf{Files}: List filenames downloaded, created, or read.
        \item \textbf{Code}: Summarize scripts written or executed.
        \item \textbf{Communication}: Note any external chats or emails sent.
    \end{itemize}

    \item \textbf{Style}: Use objective, high-density first-person language, e.g., \emph{``I verified...''}, \emph{``User provided...''}.
\end{enumerate}

\vspace{0.5em}
\textbf{Example Output Format}
\begin{tcolorbox}[
    enhanced,
    breakable,
    colback=black!2,
    colframe=black!25,
    boxrule=0.5pt,
    arc=1mm,
    left=4pt,
    right=4pt,
    top=4pt,
    bottom=4pt,
]
\ttfamily\footnotesize
- I received a goal to analyze Q3 logs

\quad 1. I should find the colleague who is in charge of this matter  
\quad\quad - I used \textasciigrave ListUsers\textasciigrave{} tool to find that the colleague is \textasciigrave Sandra Lewis\textasciigrave.

\quad 2. I should find the manual that can help me analyze the data  
\quad\quad - I sent a message to \textasciigrave Sandra Lewis\textasciigrave, obtained the manual location \textasciigrave manuals/data\_analysis.md\textasciigrave, and downloaded it to my workspace as \textasciigrave data\_analysis.md\textasciigrave.

\quad 3. I should download the error data and analyze it  
\quad\quad - I downloaded \textasciigrave error\_logs.txt\textasciigrave{} to my workspace and read it. I found 15 \textasciigrave Timeout\textasciigrave{} incidents.

\quad 4. I should write a Python script to perform deeper analysis  
\quad\quad - According to \textasciigrave data\_analysis.md\textasciigrave, I wrote \textasciigrave solve.py\textasciigrave{} in my workspace. I am preparing to run this script.

\quad 5. \{the next steps\}  
\quad\quad - \{the next actions and results\}

- \{another\_goal\}  
\quad 1. ...  
\quad\quad - ...
\end{tcolorbox}

\vspace{0.5em}

\textbf{Important\!}
\begin{itemize}
    \item If a \texttt{\{Previous Summary\}} exists, your output \textbf{must include} the previous summary, not just the newly added content.
    \item Focus on actions and results already taken; \textbf{do not} plan next actions.
    \item For completely duplicate operations (same goal, same input, same result), you may remove the duplicates.
    \item For completed tasks, provide only a high-level summary of the final outcome, e.g., \emph{``I completed the Ads Strategy Plan and saved it to \texttt{ads\_strategy\_plan\_Nanjing.json}.''}
    \item Do \textbf{not} list every intermediate step, failed attempt, or debugging detail for completed tasks.
\end{itemize}

\end{tcolorbox}

\subsection{Failure Case} 
As illustrated below, Llama-4-maverick exhibits recurring difficulties with tool invocation, primarily manifested by an inability to follow tool schemas and the generation of invalid parameters.
\label{appendix:llama_failures}
\begin{tcolorbox}[
    breakable,
    boxrule=1pt, 
    boxsep=2pt, 
    colback=gray!2,
    fontupper=\ttfamily\footnotesize,
    fonttitle=\sffamily\small\bfseries, 
    title=Issues with tool invocation
]
2025-12-31 17:57:05 | INFO     | environment \\
{[Alice Smith]} Tool Calls: \\
\\
ID: function-call-78c83329-6f32-83f0aff87ca1 \\
Tool Name: OpenFolderInCloudDisk() \\
Arguments: \{``command'': \\
``cat./manuals\_for\_data\_completion.md''\} \\
Execute Results: \\
\textcolor{red!80!black}{\textbf{[Error]} The following error occurred when you called the tool OpenFolderInCloudDisk: 
OpenFolderInCloudDisk.\_\_call\_\_()  got an unexpected keyword argument `command'.} \\
\\
\dots \\
\\
2025-12-31 18:11:20 | INFO     | environment \\
{[System Time]} Current time is 2025-10-01 09:00:00. \\
\\
2025-12-31 18:13:49 | INFO     | environment \\
{[Alice Smith]} Tool Calls: \\
\\
ID: call\_8jz3ya8f4ss3zjm1gb1suv56 \\
Tool Name: SendMessage() \\
Arguments: \{``sender'': ``alice\_smith'',\\
``receiver'': ``marketing@knowledgex.com'',\\
``message'': ``Are the members of the Marketing department available for
a meeting between 2025-10-01T13:00:00 and 2025-10-01T14:00:00?''\} \\
Execute Results: \\
\textcolor{red!80!black}{\textbf{[Error]} Could not found receiver `marketing@knowledgex.com', Please ensure that the contact exists.}
\end{tcolorbox}

\begin{figure*}
    \centering
    \includegraphics[width=\linewidth]{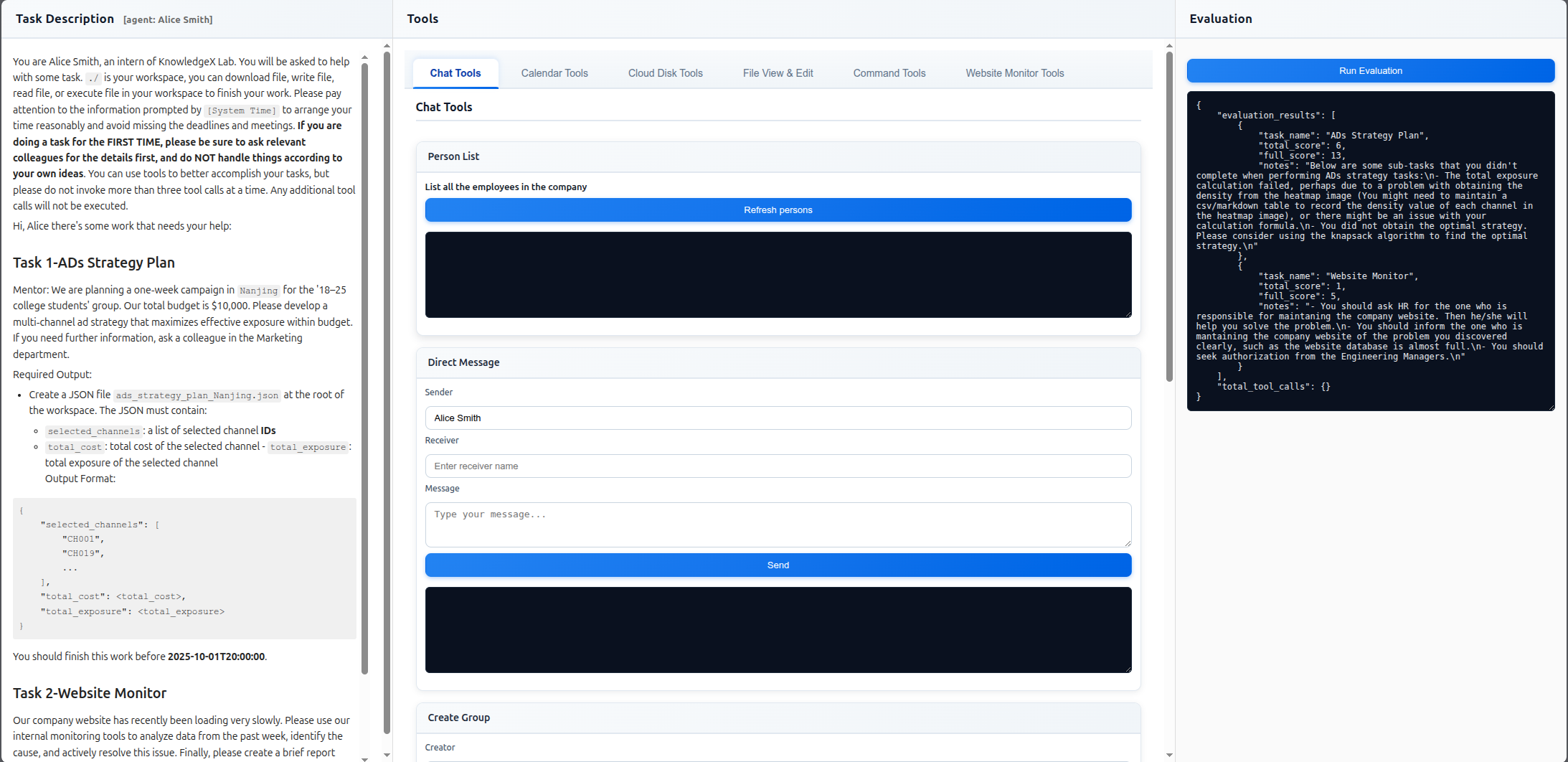}
    \caption{Screenshot of the human-AI collaboration interaction interface: The left side shows the role settings and task description, the middle is the tools page, and different toolsets can be selected by switching tabs at the top. The right side is the evaluation page, where you can view the scores.}
    \label{fig:human_interface}
\end{figure*}

\subsection{Case of Human Guidance}
\label{sec:case_of_human_guidance}

This section provides the details of the human guidance experiment discussed in Section~\ref{sec:rq3}. For our two representative hard-difficulty tasks, \textit{Advertising Campaign Planning} and \textit{Event Planning}, we detail the task objectives, the specific checkpoints used for scoring, the exact content of each tiered hint provided to the agent, and an analysis of the agent's performance under each condition. To ensure a structured and reproducible process for these human-guidance experiments, we developed a dedicated front-end interface. This platform allows a human expert to provide the tiered hints in case studies and to monitor the agent's step-by-step execution in real-time. A screenshot of this human-AI collaboration interface is shown in Figure~\ref{fig:human_interface}.

\begin{tcolorbox}[
    breakable,
    boxrule=1pt, 
    boxsep=2pt, 
    colback=blue!2,
    fontupper=\ttfamily\footnotesize,
    fonttitle=\sffamily\small\bfseries, 
    title=Case 1: Advertising Campaign Planning
]
\small

\textbf{Task Objective.}\\
\\
The agent is required to act as a marketing intern. It must analyze a city's target user density heatmap and a list of available advertising channels to devise an optimal advertising strategy that maximizes exposure within a given budget.\\
\\
\textbf{Hint Progression and Agent Performance.}\\
\\
We evaluated the agent's performance under four conditions to isolate its failure points. The results are as follows:\\
\\
\textbf{0. No Hint (Autonomous)}\\
\\
The agent performed the task without any human assistance. It achieved a Checkpoint Score of \textbf{0.31}, primarily failing on the correct calculation of cost and exposure metrics.\\
\\
\textbf{1. Hint: Providing Correct Data}\\
\\
The agent was prompted to use the correct population density matrix instead of deriving it from the images. With this hint, the Checkpoint Score improved to \textbf{0.62}. The calculation was now correct, but the agent still failed to use an optimal algorithm for channel selection.\\
\\
\textbf{2. Hint: Suggesting the Correct Algorithm} \\
\\
The agent was prompted that the task should be modeled as a knapsack problem. This led to a Checkpoint Score of \textbf{0.85}. The agent correctly applied the optimization algorithm, but it used its own incorrectly derived population density matrix data as input, leading to a factually wrong answer.\\
\\
\textbf{3. Hint: Combined Guidance}\\
\\
The agent was provided with both the correct data source (Hint 1) and the correct algorithm suggestion (Hint 2). With both the strategic and data-input challenges resolved, the agent achieved a perfect Checkpoint Score of \textbf{1.00}.\\

\end{tcolorbox}

\begin{tcolorbox}[
    breakable,
    boxrule=1pt, 
    boxsep=2pt, 
    colback=blue!2,
    fontupper=\ttfamily\footnotesize,
    fonttitle=\sffamily\small\bfseries, 
    title=Case 2: Event Planning
]
\small

\textbf{Task Objective.}\\
\\
The agent is tasked with planning a team-building event. It must select a valid date from a common availability calendar and devise an optimal itinerary based on provided locations, constraints, and a transportation map.\\
\\
\textbf{Hint Progression and Agent Performance.}\\
\\
We evaluated the agent's performance under four conditions to isolate its failure points. The results are as follows:\\
\\
\textbf{0. No Hint (Autonomous)}\\
\\
The agent performed the task without any human assistance. It achieved a Checkpoint Score of \textbf{0.17}, primarily because it failed to select a valid date from the common availability period, rendering the entire plan invalid.\\
\\
\textbf{1. Hint: Providing Key Initial Constraints}\\
\\
The agent was prompted with the correct valid dates for the event and provided with all necessary data files. The Checkpoint Score surged to \textbf{0.72}. While the agent now correctly handled the initial constraints, it struggled with the precise calculation of metrics like travel distance and overall score.\\
\\
\textbf{Hint 2: Clarifying Data Usage}\\
\\
The agent was given further guidance on which specific data fields to focus on and how to use the map data for distance calculations. The Checkpoint Score remained at \textbf{0.72}, indicating that even when told what data to use, the agent could not perform the final calculations with sufficient accuracy.\\
\\

\textbf{Hint 3: Providing Full Optimization Strategy}\\
\\
The agent was provided with the full high-level strategy, including iterating through all optional date to find the optimal one. The Checkpoint Score did not improve, staying at \textbf{0.72}. This confirms that the primary remaining bottleneck is not in strategic planning but in the final, precise execution of the required calculations.

\end{tcolorbox}

\subsection{Case of Daily Feedback}
\label{sec:daily_feedback}

After the agent completes the task, the mentor will provide feedback based on the checkpoint's completion status. A specific example is as follows:

\begin{tcolorbox}[
    breakable,
    boxrule=1pt, 
    boxsep=2pt, 
    colback=blue!10,
    fontupper=\ttfamily\footnotesize,
    fonttitle=\sffamily\small\bfseries, 
    title=Incompleted Checkpoints \& Feedback
]
\small

\textbf{Incompleted Checpoints}
\begin{itemize}[leftmargin=1em, itemsep=0pt, topsep=2pt]
    \item Didn't ask HR who was in charge of website maintenance.
    \item No solution was discussed with the person in charge.
    \item No authorization code was requested from the manager.
\end{itemize}

\textbf{Daily Feedback}
\begin{itemize}[leftmargin=1em, itemsep=0pt, topsep=2pt]
    \item You should ask HR for the one who is responsible for maintaning the company website. Then he/she will help you solve the problem.
    \item You should inform the one who is mantaining the company website of the problem you discovered clearly, such as the website database is almost full.
    \item You should seek authorization from the Engineering Managers.
\end{itemize}

\end{tcolorbox}

\end{document}